\documentclass{article}
\usepackage{iclr, times}
\usepackage{marginnote}
\usepackage{enumitem}
\usepackage[utf8]{inputenc} 
\usepackage[T1]{fontenc}    
\usepackage{url}            
\usepackage{booktabs}       
\usepackage{amsfonts}       
\usepackage{nicefrac}       
\usepackage{microtype}      
\usepackage{url,hyphenat}
\usepackage{caption,subfigure,graphicx,epstopdf,multirow,multicol,booktabs,verbatim,wrapfig,bm}
\usepackage{titlesec}
\usepackage{makecell}
\usepackage{amsthm,amssymb,amsfonts,amsmath}
\usepackage{algorithmic}
\usepackage{threeparttable}
\usepackage{listings}
\usepackage{pythonhighlight}
\usepackage{array}
\usepackage{bbm}
\usepackage{dsfont}

\usepackage{booktabs}

\usepackage[colorlinks,
            linkcolor=red,
            anchorcolor=blue,
            citecolor=purple,
            ]{hyperref}

\usepackage[vlined,commentsnumbered,linesnumbered,ruled]{algorithm2e}

\newtheorem{definition}{Definition}

\newtheorem{theorem}{Theorem}
\newtheorem{lemma}{Lemma}

\newcommand{\chameleon}{\texttt{Chameleon}\xspace}
\newcommand{\squirrel}{\texttt{Squirrel}\xspace}
\newcommand{\actor}{\texttt{Actor}\xspace}
\newcommand{\cora}{\texttt{Cora}\xspace}
\newcommand{\citeseer}{\texttt{Citeseer}\xspace}
\newcommand{\pubmed}{\texttt{Pubmed}\xspace}
\newcommand{\computer}{\texttt{Amazon-Computer}\xspace}
\newcommand{\photo}{\texttt{Amazon-Photo}\xspace}
\newcommand{\cs}{\texttt{Coauthor-CS}\xspace}
\newcommand{\physics}{\texttt{Coauthor-Physics}\xspace}
\newcommand{\computera}{\texttt{Computer}\xspace}
\newcommand{\photoa}{\texttt{Photo}\xspace}
\newcommand{\csa}{\texttt{CS}\xspace}
\newcommand{\physicsa}{\texttt{Physics}\xspace}
\newcommand{\arxiv}{\texttt{Ogbn-Arxiv}\xspace}

\title{Localized Graph Contrastive Learning}
\iclrfinalcopy %

\author{Hengrui Zhang\\
{\small University of Illinois, Chicago}\\
\texttt{\small hzhan55@uic.edu}
\And
Qitian Wu\\
{\small Shanghai Jiao Tong University}\\
\texttt{\small echo740@sjtu.edu.cn}
\And
Yu Wang\\
{\small University of Illinois, Chicago}\\
\texttt{\small ywang617@uic.edu}
\AND
Shaofeng Zhang\\
{\small Shanghai Jiao Tong University}\\
\texttt{\small sherrylone@sjtu.edu.cn\;\;\;}
\And
Junchi Yan\\
{\small Shanghai Jiao Tong University}\\
\texttt{\small yanjunchi@sjtu.edu.cn}
\And
Philip S. Yu\thanks{Corresponding author.}\\
{\small University of Illinois, Chicago}\\
\texttt{\small psyu@uic.edu}
}

\begin{document}
\maketitle

\newcommand{\zhang}[1]{{\color{red} [zhang: #1]}}
\newcommand{\themodel}{{\sc{Local-GCL}}\xspace}
\newcommand{\model}{Local-GCL}

\maketitle

\begin{abstract}
Contrastive learning methods based on InfoNCE loss are popular in node representation learning tasks on graph-structured data. However, its reliance on data augmentation and its quadratic computational complexity might lead to inconsistency and inefficiency problems. To mitigate these limitations, in this paper, we introduce a simple yet effective contrastive model named Localized Graph Contrastive Learning (\themodel in short). \themodel consists of two key designs: 1) We fabricate the positive examples for each node directly using its first-order neighbors, which frees our method from the reliance on carefully-designed graph augmentations; 2) To improve the efficiency of contrastive learning on graphs, we devise a kernelized contrastive loss, which could be approximately computed in linear time and space complexity with respect to the graph size. We provide theoretical analysis to justify the effectiveness and rationality of the proposed methods. Experiments on various datasets with different scales and properties demonstrate that in spite of its simplicity, \themodel achieves quite competitive performance in self-supervised node representation learning tasks on graphs with various scales and properties.
\end{abstract}

\section{Introduction}
Self-supervised learning has achieved remarkable success in learning informative representations without using costly handcrafted labels~\citep{cpc, bert, uncovering, moco, simclr, byol, cca-ssg, gao2021simcse}. Among current self-supervised learning paradigms, InfoNCE loss~\citep{cpc} based multi-view contrastive methods~\citep{moco, simclr, gao2021simcse} are recognized as the most widely adopted ones, due to their solid theoretical foundations and strong empirical results. Generally, contrastive learning aims at maximizing the agreement between the latent representations of two views (e.g. through data augmentation) from the same input, which essentially maximizes the mutual information between the two representations~\citep{poole2019variational}. Inheriting the spirits of contrastive learning on vision tasks, similar methods have been developed to deal with graphs and bring up promising results on common node-level classification benchmarks~\citep{dgi, mvgrl, grace, grace-ad}.

The challenge, however, is that prevailing contrastive learning methods rely on predefined augmentation techniques for generating positive pairs as informative training supervision. Unlike grid-structured data (e.g., images or sequences), it is non-trivial to define well-posed augmentation approaches for graph-structured data~\cite{grace-ad, cca-ssg}. The common practice adopted by current methods resorts to random perturbation on input node features and graph structures~\citep{graphcl}, which might unexpectedly violate the underlying data generation and change the semantic information~\citep{afgrl}. Such an issue plays as a bottleneck limiting the practical efficacy of contrastive methods on graphs. Apart from this, the InfoNCE loss function computes all-pair distance for in-batch nodes as negative pairs for contrasting signals~\citep{grace, grace-ad}, which induces quadratic memory and time complexity with respect to the batch size. Given that the model is preferred to be trained in a full-graph manner (i.e., batch size = graph size) since the graph structure information might be partially lost through mini-batch partition, such a nature heavily constrains contrastive methods for scaling to large graphs.

Some recent works seek negative-sample-free methods to resolve the scalability issue by harnessing asymmetric structures~\citep{bgrl} or feature-level decorrelation objectives~\citep{cca-ssg}. However, these methods either lack enough theoretical justification~\citep{bgrl} or necessitate strong assumptions on the data distributions~\citep{cca-ssg}. Moreover, they still require data augmentations to generate two views of the input graph, albeit non-contrastive and free from negative sampling. Some other works construct positive examples using the target's $k$-nearest-neighbors (kNN) in the latent space~\citep{nnclr, msf, afgrl}. Nonetheless, the computation of nearest neighbors could be cumbersome, time-consuming, and, therefore hard to scale.

\textbf{Presented Work.} To cope with the dilemmas above, in this paper we introduce \textbf{\underline{Local}}ized \textbf{\underline{G}}raph \textbf{\underline{C}}ontrastive \textbf{\underline{L}}earning (\themodel in abbreviation), a light and \textbf{augmentation-free} contrastive model for self-supervised node-level representation learning on graphs. \themodel benefits from two key designs. \textbf{First}, it does not rely on data augmentation to construct positive pairs. Instead, inspired by the graph homophiliy theory~\citep{homo}, it directly treats the first-order neighboring nodes as the positive examples of the target node. This not only increases the number of positive examples for each node but also helps our model get rid of complicated data augmentations. Besides, the computation of positive pairs can be performed in linear time and space complexity w.r.t the graph size, bringing no additional cost for the model. \textbf{Second}, to deal with the quadratic complexity curse of contrastive loss (i.e., InfoNCE loss~\citep{cpc}), we propose a surrogate loss function in place of the negative loss term in the vanilla InfoNCE loss, which could be efficiently and accurately approximated within linear time and space complexity~\citep{rff, rff-survey, orf}. Such a design greatly improves the efficiency of our model.

We evaluate the proposed methods on seven public node classification benchmarks with various scales. The empirical results demonstrate that though not using any graph augmentations, our method achieves state-of-the-art performance on six of seven datasets. On the challenging \arxiv dataset, our method can also give a competitive performance with a much training speed compared with other scalable models. Experiments on three heterophily graphs demonstrate that besides homophily graphs, \themodel can also perform well on graphs with low homophily ratios.

\noindent\textbf{We summarize the highlights of this paper as follows:}
\begin{itemize}[topsep=0in,leftmargin=0em,wide=0em]
\item[\textbf{1)}]  We introduce \themodel, a simple model for contrastive learning on graphs, where the positive example is fabricated using the first-order neighborhood of each node. This successfully frees node-level contrastive learning methods from unjustified graph augmentations.

\item[\textbf{2)}] To overcome the quadratic complexity curse of contrastive learning, we propose a kernelized contrastive loss computation that can precisely approximate the original loss function within linear complexity w.r.t. graph size. This significantly reduces the training time and memory cost of contrastive learning on large graphs.

\item[\textbf{3)}] Experimental results show that without data augmentation and other cumbersome designs \themodel achieves quite competitive results on a variety of graphs of different scales and properties. Furthermore, \themodel demonstrates a better balance of model performance and efficiency than other self-supervised methods.
\end{itemize}

\section{Background and Related Works}
\subsection{Contrastive Representation Learning}
Inspired by the great success of contrastive methods in learning image representations~\citep{cpc, dim, cmc, moco, simclr}, recent endeavors develop similar strategies for node-level tasks in graph domain~\citep{dgi, mvgrl, grace, grace-ad}. Among graph contrastive learning methods, the most popular methods should be those based on the InfoNCE loss~\citep{cpc} due to their simple concepts and better empirical performance. InfoNCE-based graph contrastive learning methods, including GRACE~\citep{grace} and GCA~\citep{grace-ad} aim to maximize the similarity of positive node-node (or graph-graph) pairs (e.g., two views generated via data augmentation) and minimize the similarity of negative ones (e.g., other nodes/graphs within the current batch). However, they require well-designated data augmentations that {could positively inform} downstream tasks~\citep{mini-1}. The quadratic complexity also limits their applications to larger batch sizes/datasets.

\subsection{Augmentation-Free Self-supervised Learning on Graphs}
Besides graph contrastive learning~\citep{grace, grace-ad} and the similar two-branched models~\citep{bgrl, cca-ssg} that use graph augmentations to create positive pairs, there is another line of work exploring the rich structure information in the graph to create self-supervised signals. Graph AutoEncoders~\citep{gae}, for example, learn node embeddings unsupervisedly through reconstructing the adjacency matrix, while GMI~\citep{gmi} maximizes the mutual information of both features and edges between inputs and outputs. Inspired by the success of contrastive learning, some recent work explores constructing positive pairs using reliable graph information instead of data augmentations. For example, SUGRL~\citep{sugrl} employs two encoder models: one is a GCN, and the another is an MLP, to generate two sets of node embedding from different sources. Then for each node, the positive pair can be constructed using its GCN output and MLP output. AFGRL~\citep{afgrl} and AF-GCL~\citep{afgcl-guarantee} treat nodes in the target node's multi-hop neighborhood as candidate positive examples and use well-designed similarity measures to select the most similar nodes as positive examples. \themodel, by contrast, treats all neighbors equally without discrimination. This not only simplifies the model design but also offers a theoretical guarantee for the effectiveness of the proposed model. Empirical results also demonstrate that our method can achieve better performance than theirs. 
\subsection{Random Fourier Features}\label{sec-intro-rff}
Though InfoNCE-loss-based based methods are successful in self-supervised node representation learning, their quadratic time and memory complexity with respect to the graph size prevent them from being applied to graphs with tens of thousands of nodes. This paper seeks to address this issue by optimizing a surrogate loss function with Random Fourier Features (RFF or Random Features in short)~\citep{rff, rff-survey}. RFF is an effective technique for enhancing the scalability of kernel methods such as SVM, ridge regression~\citep{ridge_regression} and independence test~\citep{kernel_dependence, SSL-HSIC}. Also, recently, RFF has been adopted to develop linear Transformers by approximating the softmax attention~\citep{performer, rfa}. Given $d$-dimensional vectors $\bm{x}$ and $\bm{y}$ and a shift-invariant kernel $\kappa(\cdot)$, RFF constructs an explicit mapping $\psi$: $ \mathbb{R}^{d} \rightarrow \mathbb{R}^D$, such that $\kappa(\bm{x},\bm{y}) \approx \psi(\bm{x})^{\top}\psi(\bm{y})$, which reduces the quadratic computation cost of the kernel matrix to a linear one w.r.t data size. Generally, given a positive definite shift-invariant kernel $\kappa(\bm{x}, \bm{y}) = f(\bm{x} - \bm{y}) $, the Fourier transform $p$ of kernel $\kappa$ is $p(\bm{\omega}) = \frac{1}{2\pi} \int e^{-j\bm{\omega}'\Delta}k(\Delta)\mathrm{d}\Delta$, where $\Delta = \bm{x} - \bm{y}$. Then we could draw $D$ i.i.d. samples $\bm{\omega}_1, \cdots, \bm{\omega}_D \in \mathbb R^{d}$ from $p$, and $\psi(\bm{x})$ is given by:
\begin{equation}\label{eqn-rff1}
\begin{aligned}
    \psi(\bm{x}) = \frac{\left[\cos(\bm{\omega}_1^{\top}\bm{x}), \cdots, \cos(\bm{\omega}_D^{\top}\bm{x}), \sin(\bm{\omega}_1^{\top}\bm{x}), \cdots, \sin(\bm{\omega}_D^{\top}\bm{x})\right]^{\top}}{\sqrt{D}}.
\end{aligned}
\end{equation}
Let $\bm{W} = [\bm{\omega}_1, \cdots, \bm{\omega}_D] \in \mathbb{R}^{d\times D}$ be a linear transformation matrix, one may realize that the computation of Eq.~\ref{eqn-rff1} entails computing $\bm{W}^{\top}\bm{x}$. Specifically, when $\kappa(\cdot)$ is a standard Gaussian kernel (a.k.a., RBF kernel), each entry of $\bm{W}$ can be directly sampled from a standard Gaussian distribution. The improved variants of RFF mainly concentrate on different ways to build the transformation matrix $\bm{W}$, so as to further reduce the computational cost~\citep{fastfood} or lower the approximation variance~\citep{orf}.

\section{Contrastive Learning on Graphs with InfoNCE Loss}\label{sec:background}
The InfoNCE~\citep{cpc} loss 
has been employed in various areas and has shown great power in learning informative representations in a self-supervised manner. Given the embedding of a target data point $\bm{z}_{i}$, together with one positive embedding $\bm{z}_{i}^{+}$ and a set of negative embeddings $\{\bm{z}_{j}^{-}\}_{j=1}^{M-1}$, the InfoNCE loss aims at discriminating the positive pair $(\bm{z}_i, \bm{z}_i^{+})$ from the negative pairs $\{(\bm{z}_i, \bm{z}_j^{-}) \}_{j=1}^{M-1} $ via the following contrastive objective function:
\begin{equation}\label{eqn-infonce}
    \mathcal{L}_{\rm{InfoNCE}}(i) = -\log \frac{\exp({f(\bm{z}_i, \bm{z}_i^{+}) / \tau})}{\exp({f(\bm{z}_i, \bm{z}_i^{+}) / \tau})+\sum\limits_{j=1}^{M-1} \exp({f(\bm{z}_i^+, \bm{z}_j^{-})}/\tau)},
\end{equation}
where $f(\cdot, \cdot)$ is a similarity measure, usually implemented as the simple dot product $f(\bm{x}, \bm{y}) = \bm{x}^{\top}\bm{y}$~\citep{moco, simclr}, and $\tau$ is the temperature hyperparameter. Note that $\bm{z}_i, \bm{z}_i^{+}$ and $\bm{z}_i^{-}$ are all $\bm{\ell}_2$ normalized to have a \textbf{unit norm}, i.e., $\Vert \bm{z}_i \Vert_2^2 = 1$. In the context of node-level contrastive learning, each $\bm{z}_i$ denotes the embedding of node $i$, which is the output of an encoder model which takes the node feature $\bm{x}_i$ and the graph structure as input. As demonstrated in Eq.~\ref{eqn-infonce}, the InfoNCE loss is composed of two terms: 1) the numerator term on the positive pair that maximizes the similarity between positive pairs (positive term); 2) the denominator term on $M-1$  negative pairs 
and one positive pair (for simplicity, we call all the $M$ terms in the denominator "negative terms") that encourages the embeddings of negative pairs to be distinguished (negative term). Like the study of contrastive learning on vision and language, recent node-level contrastive learning research mainly focuses on how to construct/select more informative positive and negative examples.

The most popular method for constructing \textbf{positive examples} on graphs is through data augmentation. For example, MVGRL~\citep{mvgrl} uses graph diffusion to generate a fixed, augmented view of the studied graph. GCC~\citep{gcc} resorts to random walk to sample subgraphs of the anchored node as positive examples. GRACE~\citep{grace} and GCA~\citep{grace-ad} employ random graph augmentation--\textit{feature masking} and \textit{edge removing} to generate two views of the input graph before every training step. However, recent studies have shown that existing graph augmentations might unexpectedly change or lose the semantic information~\citep{afgrl, afgcl-guarantee} of the target node, hurting the learned representation. This motivates us to investigate the possibility of constructing positive examples without using random, ambiguous graph augmentations.

Compared with the construction of positive examples, \textbf{negative examples} are much easier to select. A common adopted and effective practice is to use all other nodes in the graph as negative examples~\citep{grace, grace-ad, cca-ssg}. However, the computation of the negative term of InfoNCE has $\mathcal{O}(|\mathcal{V}|^2d)$ time complexity and $\mathcal{O}(|\mathcal{V}|^2)$ memory complexity, where $|\mathcal{V}|$ is the number of nodes and $d$ is the embedding dimension. Note that the real-world graphs are usually very large, Eq.~\ref{eqn-infonce} can hardly be used in a full-graph training manner. One plausible remedy is to use down-sampling to zoom in on a fraction of negative samples for once feedforward computation. However, theoretical analysis demonstrates InfoNCE loss benefits from a large number of negative examples~\citep{poole2019variational}, and empirical observations show that reducing the number of negative examples leads to worse performance on downstream tasks~\citep{bgrl} (also see our experiments in Sec.~\ref{sec:exp-scale}). As a result, it would be promising to devise an efficient and scalable computation method for the contrastive loss above, without reducing the number of negative examples.

In the next section, we propose Localized Graph Contrastive Learning with Kernel Approximation (\themodel in short) to mitigate the above issues. \themodel distinguishes itself from previous contrastive methods in the following two aspects: 1) instead of constructing positive examples for target nodes using data augmentations or other techniques, we directly treat all the first-order neighbors of each target node as its positive examples; 2) we introduce a Gaussian kernelized surrogate loss  function, which could be accurately and efficiently approximated with linear time and memory complexity in place of the vanilla InfoNCE loss.

\section{Methodology}\label{sec:method}
\subsection{Anchoring Node Contrastive Learning on Topology}\label{sec-positive}
Contrastive learning is firstly investigated in unsupervised visual representation learning where the data points are i.i.d. sampled from the data distribution, and the ground-truth labels are unknown during training. As a result, we have to employ well-defined augmentations to create two views of the same image that are likely to share the same ground-truth label. Different from images or texts where each input data is often assumed i.i.d. generated from a certain distribution and there is no explicit prior knowledge about the relationship among different instances, in the graph domain, nodes within the same graph are highly inter-dependent given the input graph structure. That is, the graph structure could provide additional information that reflects the affinity of node pairs.
Furthermore, in network (graph) science, it is widely acknowledged that many real-world graphs (such as social networks~\citep{dansar, homo} and citation networks~\citep{homo-citation}) conform with the homophily phenomenon that similar nodes may be more likely to attach than dissimilar ones~\citep{lp, homo}.
Such a phenomenon also inspires the core principle followed by the designs of modern graph neural networks, i.e., connected nodes should have similar embeddings in the latent space~\citep{beyond-homo}. This motivates us to directly treats all the first-order neighboring nodes of the target node as positive examples. Formally, the loss for the positive pairs of node $i$ can be written as  
\begin{equation}\label{eqn:pos}
    \mathcal{L}_{\rm pos}(i) = -\log \sum\limits_{j \in \mathcal{N}(i)} \exp (\bm{z}_i^{\top}\bm{z}_j / \tau) / |\mathcal{N}(i)|.
\end{equation}
For the negative term, we can directly use all nodes in the graph as negative examples. Then the loss for the negative pairs of node $i$ can be formulated as:
\begin{equation}\label{eqn:neg}
  \mathcal{L}_{\rm neg}(i) = \log \sum\limits_{k \in \mathcal{V}} \exp (\bm{z}_i^{\top}\bm{z}_k / \tau).  
\end{equation}
The overall loss function is simply the sum of $\mathcal{L}_{\rm pos} $ and $\mathcal{L}_{\rm neg}$, then averaged over all nodes:
\begin{equation}\label{eqn:loss-model}
    \mathcal{L}_{\rm \themodel} = -\frac{1}{|\mathcal{V}|} \sum\limits_{i=1}^N  \log \frac{\sum\limits_{j \in \mathcal{N}(i)}  \exp (\bm{z}_i^{\top}\bm{z}_j/\tau) / |\mathcal{N}(i)|}{\sum\limits_{k \in \mathcal{V}} \exp (\bm{z}_i^{\top}\bm{z}_k /\tau)} .
\end{equation}
Then we'd like to provide a theoretical analysis of the effectiveness of constructing positive examples directly with the first-order neighbors. First, let's introduce some notations that will be used in the following analysis. We denote a graph by $\mathcal{G} = (\mathcal{V}, \mathcal{E})$, where $\mathcal{V}$ is the node st and $\mathcal{E}$ is the edge set. The adjacency matrix and degree matrix are denoted by $\bm{A}$ and $\bm{D}$, respectively. Let $\tilde{\bm{A}} = \bm{D}^{-1/2}\bm{A}\bm{D}^{-1/2}$ be the symmetric normalized adjacency matrix, and $\bm{L} = \bm{I} - \tilde{\bm{A}}$ be the symmetric normalized graph Laplacian matrix. Denote the eigenvalues of $\bm{L}$ in Zn ascending order by $\{\lambda_i\}_{i=1}^{|\mathcal{V}|}$. Finally we denote the node embedding matrix $\bm{Z} = \{ \bm{z}_i \}_{i=1}^{|\mathcal{V}|} \in \mathbb{R}^{|\mathcal{V}|\times d} $ and the one-hot label matrix by $\bm{Y} = \{ \bm{y}_i \}_{i=1}^{|\mathcal{V}|} \in \mathbb{R}^{|\mathcal{V}|\times c}$. Without loss of generality we assume $d \le |\mathcal{V}|$.

We then give a formal definition of graph homophily ratio $\phi$:
\begin{definition}\label{def:homo_ratio}
(Graph Homophily Ratio)
For a graph $\mathcal{G} = (\mathcal{V}, \mathcal{E})$ with adjacency matrix $\bm{A}$,  its  homophily ratio $\phi$ is defined as the probability that two connected nodes share the same label:
\begin{equation}\label{eqn:homo-ratio}
    \phi = \frac{\sum_{i,j \in \mathcal{V}} A_{ij} \cdot \mathbbm{1}[{\bm{y}_i = \bm{y}_{j}}]}{\sum_{i,j \in \mathcal{V}} A_{ij}} = \frac{\sum_{i,j \in \mathcal{V}} A_{ij} \cdot \mathbbm{1}[{\bm{y}_i = \bm{y}_{j}}]}{|\mathcal{E}|}
\end{equation}
\end{definition}

 With the conclusions in \citet{contrastive-spectral} and \citet{contrastive-guarantee}, which build connections between contrastive loss and spectral method, the following theorem guarantees the linear classification error of the embeddings learned from \themodel:
\begin{theorem}\label{theorem:error}
Let $\bm{Z}^*$ be the global minimizer of  Eq.~\ref{eqn:loss-model}, then for any labeling function $\hat{y}: \mathbb{V} \rightarrow \mathbb{R}^{c}$ with graph homophily $\phi$, there exists a linear classifier $\bm{B}^* \in \mathbb{R}^{d \times c}$ with norm $\Vert \bm{B}^* \Vert_F \le 1/(1-\lambda_d) $ such that
\begin{equation}\label{eqn:error}
    \mathbb{E}_{i \in \mathcal{V}} \left[\Vert \hat{\mathop{y}}(i)  - {\bm{B}^*}^{\top}\bm{z}_i^{*} \Vert_2^2 \right] \le \frac{1-\phi}{\lambda_{d+1}}
\end{equation}
\end{theorem}
See proof in Appendix~\ref{appendix-proof-error}. Theorem~\ref{theorem:error} demonstrates that the linear classification accuracy of the learned embeddings through \themodel is bounded by the homophily ratio $\phi$ and the $d+1$ smallest eigenvalue $\lambda_{d+1}$. Specifically, the larger the homophily ratio, the smaller the prediction error. Besides, Eq.~\ref{eqn:error} indicates that a larger embedding dimension can lead to better classification accuracy, which is also validated empirically in Sec.~\ref{sec-experiments}.

\subsection{Fast and Efficient Approximation for the Negative Loss}\label{sec-negative}
We next probe into how to reduce the computation complexity of the negative loss term $\mathcal{L}_{\rm neg} = \sum\limits_{i \in \mathcal{V}} \mathcal{L}_{\rm neg} (i) $, for which the $|\mathcal{V}|^2$ pair-wise similarities can be efficiently and precisely approximated with efficient computation methods for kernel functions~\citep{rff, rff-survey}. 

We notice that the pair-wise similarity (in dot-product format) in Eq.~\ref{eqn:neg} is essentially a Gaussian kernel function, i.e.,
\begin{equation}
\exp(\bm{z}_i^{\top}\bm{z}_j/\tau)  = \exp(\frac{2-\Vert \bm{z}_i - \bm{z}_j\Vert_2^2}{2\tau}) = e* \exp(\frac{-\Vert \bm{z}_i - \bm{z}_j \Vert}{2\tau}) = e\cdot\kappa^{G}(\bm{z}_i, \bm{z}_k ; \sqrt{\tau)},
\end{equation}
where $\kappa^{G}(\bm{z}_i, \bm{z}_k ; \sqrt{\tau)}$ is the Gaussian kernel function with bandwidth $\sqrt{\tau}$. This motivates us to seek for a projection function ${\psi}(\bm{x}) (\mathbb{R}^{d} \rightarrow \mathbb{R}^{2D})$ such that $\kappa^{G}(\bm{x}, \bm{y};\sqrt{\tau}) \approx {\psi}(\bm{x})^{\top}{\psi}(\bm{y})$. Then we are able to use a surrogate negative loss function as a remedy of the previous one:
\begin{equation}\label{eqn-decompose}
    \mathcal{L}_{{neg}} \triangleq \frac{1}{|\mathcal{V}|} \sum_{i \in \mathcal{V}} \left(\log {\sum_{j\in\mathcal{V} } {\psi}(\bm{h}_i)^{\top} {\psi}(\bm{h}_j)}\right) =  \frac{1}{|\mathcal{V}|} \sum_{i\in \mathcal{V}} \left( \log \left({ {\psi}(\bm{h}_i)^{\top} \sum_{j\in \mathcal{V} } {\psi}(\bm{h}_j)}\right) \right).
\end{equation}
Once we obtain all the projected vectors $\{\bm{\psi}(\bm{h}_{i})\}_{i=1}^{|\mathcal{V}|}$, the summation term $\sum_{j \in \mathcal{V}} {\psi}(\bm{h}_j)$ in Eq.~\ref{eqn-decompose} could be calculated within $\mathcal{O}(|\mathcal{V}|D)$ in advance (where $D$ is the dimension of projected vectors). In addition to the $\mathcal{O}(|\mathcal{V}|D)$ cost for computing the loss, the overall computational cost will be $\mathcal{O}(|\mathcal{V}|D + |\mathcal{V}|D) = \mathcal{O}(VD)$, which is linear to the number of nodes. Then we discuss how to formulate the projection function ${\psi}(\cdot)$ below.

\paragraph{Linear-Order Projection}\label{sec-rff}
The theory of Random Fourier Features~\citep{rff} introduced in Sec.~\ref{sec-intro-rff} demonstrates that a Gaussian kernel function $\kappa^{G}(\bm{x},\bm{y} ; \sqrt{\tau})$ could be unbiasedly estimated with $\psi(\bm{x})^{\top}{\psi}(\bm{y})$, and the projection function $\psi(\bm{x})$ is defined as follows:
\begin{equation}\label{eqn-rff2}
    \psi(\bm{x}) = \frac{[\cos(\bm{W}^{\top}\bm{x}), \sin(\bm{W}^{\top}\bm{x})]}{\sqrt{D}} = \frac{\left[\cos(\bm{\omega}_1^{\top}\bm{x}), \cdots, \cos(\bm{\omega}_D^{\top}\bm{x}), \sin(\bm{\omega}_1^{\top}\bm{x}), \cdots, \sin(\bm{\omega}_D^{\top}\bm{x})\right]^{\top}}{\sqrt{D}}.
\end{equation}
where $\bm{W} = [\bm{\omega}_1^{\top}, \cdots, \bm{\omega}_D^{\top}]$, and each $\bm{\omega}_i$ is sampled from the Gaussian distribution $p(\bm{\omega}) = \mathcal{N} (\bm{0}, \tau\bm{I})$. $D$ is the number of total samples. Usually, the larger the sampling number $D$ is, the more accurate the approximation will be:
\begin{theorem}\label{theorem-rff}
Let $\{\bm{\omega}_i\}_{i=1}^{D}$ be i.i.d samples from Gaussian distribution $\mathcal{N}(\bm{0}, \tau\bm{I})$, and $ \psi(\bm{x}) $ is given by Eq.~\ref{eqn-rff2}, then with probability at least $1-\varepsilon$, the approximation error $\Delta = |{\psi}(\bm{h}_i)^{\top} {\psi}(\bm{h}_j) - \kappa^{G} (\bm{h}_i, \bm{h}_j) | $ will be bounded by $\mathcal{O}\left({\frac{1-\exp(-4/\tau)}{\sqrt{2D\varepsilon}}}\right)$.
\end{theorem}
Theorem~\ref{theorem-rff} suggests that the Gaussian kernel function could be accurately approximated with Random Fourier Features as long as we sample enough number of linear transformation vectors $\bm{\omega}$ 
(i.e., $D$ should be large enough). Note that the computation of linear projection $ \bm{W}\bm{h}_i$ ($i = 1 \in \mathcal{V}$) requires additional $\mathcal{O}(|\mathcal{V}|dD)$ time. A large number of the projection dimension $D$ makes the computation of Eq.~\ref{eqn-decompose} still expensive for high-dimensional data.

\paragraph{Log-Order Projection}\label{sec-sorf}
To handle the above issues, we resort to Structured Orthogonal Random Features (SORF)~\citep{orf}, another Random Feature technique that imposes structural and orthogonality on the linear transformation matrix $\bm{W}$. Different from the vanilla RFF which directly samples linear transformation vectors $\bm{\omega}$ form normal distribution to construct the transformation matrix $\bm{W}_{\rm rff} = [\bm{\omega}_1^{\top}, \cdots,  \bm{\omega}_D^{\top}] \in \mathbb{R}^{d\times D}$, SORF assumes $D = d$ and constructs a structured orthogonal transformation matrix through the continued product of a series of structured matrixes $
     \bm{W}_{\rm{sorf}} = \frac{\sqrt{d}}{\sigma} \bm{HD}_1\bm{HD}_{2}\bm{HD}_{3}$,
where $\bm{D}_i \in \mathbb{R}^{d\times d}, i = 1, 2, 3 $, are diagonal ``sign-ﬂipping” matrices, with each diagonal entry sampled from the Rademacher distribution, and $\bm{H} \in \mathbb{R}^{d \times d}$ is the normalized Walsh-Hadamard matrix. By this definition, the projected dimension is restricted to $d$, but could be extended to any dimension by concatenating multiple independently generated features or simply using a proportion of them. 
 The stand-out merit of SORF is that it can be computed in $\mathcal{O}(|\mathcal{V}|D\log d)$ time using fast Hadamard transformation~\citep{hadamard}, and hardly requires extra memory cost using in-place operations~\citep{orf}. This further reduces its complexity and endows our method with desirable scalability to not only larger dataset sizes but also larger embedding dimensions. If not specified, in the following section, the term \themodel denotes our method equipped with SORF to approximate the negative loss.
\section{Experiments}\label{sec-experiments}
We conduct experiments to evaluate the proposed method by answering the following questions:
\begin{itemize}
    \item \textbf{RQ1}: How does \themodel perform compared with other self-supervised learning methods on graphs with different properties?
    \item \textbf{RQ2}: How do the specific designs of \themodel, such as the embedding dimension and projection dimension, affect its performance?
    \item \textbf{RQ3}: What's the empirical memory and time consumption of \themodel compared with prior works? Is \themodel able to scale to real-world large-scale graphs with satisfying performance?
\end{itemize}

\subsection{Experimental Setups}
\paragraph{Datasets.}
We evaluate \themodel on various datasets with different scales and properties. Following prior works~\citep{grace, cca-ssg} on self-supervised node representation learning, we adopt the $7$ common small-scale benchmarking graphs: \cora, \citeseer, \pubmed, \computer, \photo, \cs, and \physics. To evaluate the performance and scalability of \themodel on larger graphs, we also adopt \arxiv with about 170k nodes, on which a lot of methods fail to scale due to memory issues. Furthermore, we adopt three widely used heterophily graphs, \chameleon, 
\squirrel, and \actor to evaluate the generalization ability of \themodel on graphs where the graph homophily assumption does not hold. The detailed introduction and statistics of these datasets are presented in Appendix~\ref{appendix-dataset}
\paragraph{Baselines.}
We consider representative prior self-supervised models for comparison. We classify previous methods into two types: 1) \textbf{Augmentation-based}, which uses data augmentations to generate positive or negative pairs. 2) \textbf{Augmentation-free}, which uses other information rather than any form of data augmentation to create self-supervised signals. For augmentation-based baselines, we consider DGI~\citep{dgi}, MVGRL~\citep{mvgrl}, GRACE~\citep{grace}, GCA~\cite{grace-ad}, BGRL~\citep{bgrl} and CCA-SSG~\citep{cca-ssg}. For augmentation-free baselines, we consider GMI~\citep{gmi}, SUGRL~\citep{sugrl}, AFGRL~\citep{afgrl}, AF-GCL~\citep{afgcl-guarantee}.
\paragraph{Evaluation Protocols.}
 We follow the linear evaluation scheme in previous works~\citep{dgi, mvgrl, grace-ad, cca-ssg}: For each dataset, i) we first train the model on all the nodes in a graph without supervision by optimizing the objective in Eq.~\ref{eqn:loss-model}; ii) after the training ends, we freeze the parameters of the encoder and obtain all the nodes' embeddings, which are subsequently fed into a linear classifier (i.e., a logistic regression model) to generate a predicted label for each node. In the second stage, only nodes in the training set are used for training the classifier, and we report the classification accuracy on testing nodes. 
\paragraph{Implementation Details.}
The model is implemented with PyTorch and DGL~\citep{dgl}. All experiments are conducted on an NVIDIA V100 GPU with 16 GB memory unless specified. We adopt \texttt{structure-net}\footnote{\url{https://github.com/HazyResearch/structured-nets}} to do fast Walsh-Hadamard transformation, which enables much faster forward and backward computation with CUDA accelerations.
We use the Adam optimizer~\citep{adam} for both self-supervised pretraining training and linear evaluation using logistic regression. Following previous works~\citep{grace-ad, cca-ssg, bgrl, afgcl-guarantee}, we the random 1:1:8 split for \computer, \photo, \cs and \computer, and use the public recommended split for the remaining datasets. For each experiment, we report the average test accuracy with the standard deviation over 20 random initialization. If not specified, we use a two-layer GCN model as the encoder to generate node embeddings. More detailed hyperparameter settings for each dataset can be found in Appendix~\ref{appendix:hyper}
\subsection{Numerical Results}\label{exp-major}

\paragraph{Results on common graphs.}
We first report the results of node classification tasks on small-scale citation networks and social networks in Table~\ref{tbl-exp-main}. We see that although not relying on data augmentations or other complicated operations to create self-supervised signal, \themodel performs competitively with our self-supervised baselines, achieving state-of-the-art performance in 6 out of 7 datasets. It is worth noting that the competitive InfoNCE-loss based contrastive methods GRACE and GCA suffer from OOM on \physics datasets due to the $\mathcal{O}(|\mathcal{V}|^2d)$ space complexity, while \themodel can avoid such an issue thanks to the linear approximation of the negative loss.

\begin{table*}[tb!]
	\centering
	\caption{Comparison of self-supervised methods on benchmarking graphs. We group each method according to whether it relies on graph augmentation.}
	\label{tbl-exp-main}
	\small
	\begin{threeparttable}
        \scalebox{0.87}
        {
		\begin{tabular}{l|lccccccc}
			\toprule[1.0pt]
		       & Methods & \cora & \citeseer & \pubmed & \computera & \photoa & \csa & \physicsa  \\
			\midrule
              \multirow{6}{*}{\rotatebox{90}{Aug-based}} & DGI & 82.3$\pm$0.6  & 71.8$\pm$0.7 & 76.8$\pm$0.6 & 83.95$\pm$0.47  & 91.61$\pm$0.22 & 92.15$\pm$0.63 & 94.51$\pm$0.52\\
                & MVGRL & 83.5$\pm$0.4  & 73.3$\pm$0.5 & 80.1$\pm$0.7 & 87.52$\pm$0.11  & 91.74$\pm$0.07 & 92.11$\pm$0.12 & 95.33$\pm$0.03  \\
                & GRACE & 81.9$\pm$0.4 & 71.2$\pm$0.5 & 80.6$\pm$0.4 & 86.25$\pm$0.25  & 92.15$\pm$0.24 & 92.93$\pm$0.01 & OOM$^*$ \\
                & GCA & 82.3$\pm$0.4 &  72.1$\pm$0.4 &  80.7$\pm$0.5 & 87.85$\pm$0.31  & 92.49$\pm$0.09 & 93.10$\pm$0.01 & OOM   \\
                & BGRL & 82.7$\pm$0.6 & 71.1$\pm$0.8  & 79.6$\pm$0.5 & 89.69$\pm$0.37 & 93.07$\pm$0.28 & 92.59$\pm$0.17 & 95.48$\pm$0.08 \\
                & CCA-SSG & 84.2$\pm$0.4 & 73.1$\pm$0.3 & {81.6}$\pm${0.4}   & {88.74}$\pm${0.28} & {93.14}$\pm${0.14} & {93.31}$\pm${0.22} & 95.38$\pm$0.06\\
                \midrule
                \multirow{5}{*}{\rotatebox{90}{Aug-free}}  & GMI & 82.4$\pm$0.6  & 71.7$\pm$0.2 & 79.3$\pm$1.0 & 84.22$\pm$0.52 & 90.73$\pm$0.24 & OOM & OOM \\
                & SUGRL & 83.4$\pm$0.5 & 73.0$\pm$0.4 & 81.9$\pm$0.3 & 
                {88.93}$\pm${0.21} & 93.07$\pm$0.15 & 92.83$\pm$0.23 & 95.38$\pm$0.11 \\
                 & AFGRL & 81.3$\pm$0.2 & 68.7$\pm$0.3 & 80.6$\pm$0.4 & 
                \textbf{89.88}$\pm$\textbf{0.33} & \textbf{93.22}$\pm$\textbf{0.28} & 93.27$\pm$0.17 & OOM   \\
                & AF-GCL & 83.2$\pm$0.2  & 72.0$\pm$0.4 & 79.1$\pm$0.8 & 
                {89.68}$\pm${0.19} & 92.49$\pm$0.31 & 91.92$\pm$0.10 & 95.12$\pm$0.15   \\
                & \themodel & \textbf{84.5}$\pm$\textbf{0.4} & \textbf{73.6}$\pm$\textbf{0.4} & \textbf{82.1}$\pm$\textbf{0.5}  & 88.81$\pm$0.37 & 
                \textbf{93.25}$\pm$\textbf{0.40} & \textbf{94.90}$\pm$\textbf{0.19}  & \textbf{96.33}$\pm$\textbf{0.13} \\
		\bottomrule[1.0pt]
		\end{tabular}
        }
		\begin{tablenotes}
        \item[$*$] OOM indicates out-of-memory on an NVIDIA-V100 GPU of 16G memory.
        \end{tablenotes}
	\end{threeparttable}
\end{table*}

\begin{figure}[t]
    \begin{minipage}[h]{0.45\linewidth}
    \vspace{0pt}
    \centering
    \captionof{table}{Performance on \arxiv dataset. As recommended, we report both the validation accuracy and test accuracy.}
    \label{tbl-exp-arxiv}
    \small
    {   
	\begin{threeparttable}
        \scalebox{1.0}
        {
	\begin{tabular}{lcc}
		\toprule[1.0pt]
             Methods & Validation & Test \\
            \midrule
            DGI  & 71.21$\pm$0.23  & 70.32$\pm$0.25    \\
            GMI & OOM & OOM \\
            MVGRL  & OOM   & OOM \\
            GRACE  & OOM   & OOM \\
            GCA & OOM   & OOM \\
            BGRL & \textbf{72.71}$\pm$\textbf{0.22} & \textbf{71.54}$\pm$\textbf{0.17}  \\
            CCA-SSG & 72.31$\pm$0.18 & 71.21$\pm$0.20 \\
            AFGRL & OOM & OOM \\
            \midrule 
            \themodel & 72.29$\pm$0.25 & 71.34$\pm$0.25 \\
            \bottomrule[1.0pt]
	\end{tabular}
		}

	\end{threeparttable}
	}
    \end{minipage}
        \begin{minipage}[h]{0.5\linewidth}
        \vspace{0pt}
        \centering
        \includegraphics[width=1.0\textwidth,angle=0]{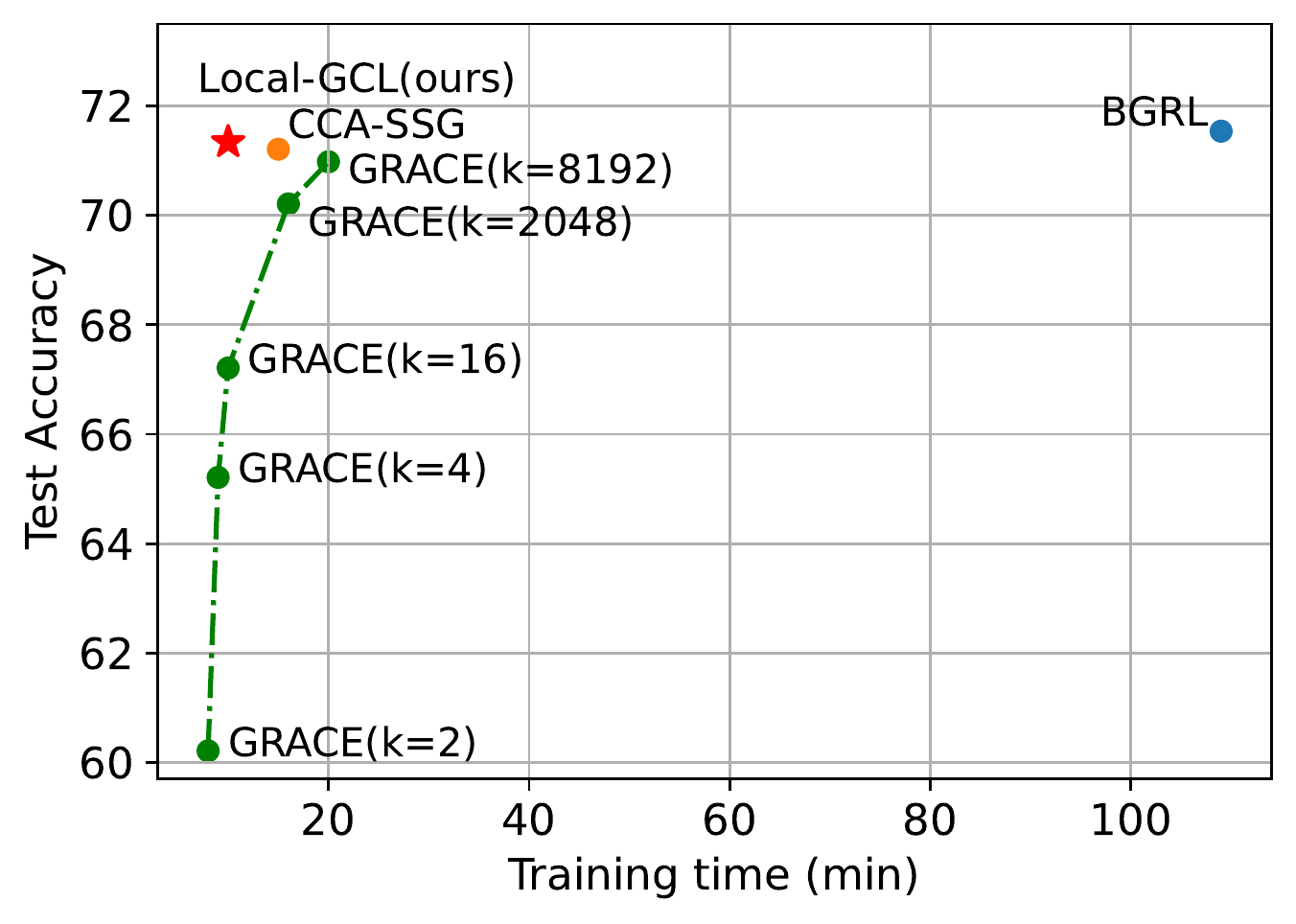}
        \vspace{-15pt}
        \caption{Test accuracy and training time comparison of \themodel, BGRL, CCA-SSG and GRACE with subsampling.}
        \label{fig:scale}
    \end{minipage}
\end{figure}
\paragraph{Results on Ogbn-Arxiv.} Then, we evaluate the effectiveness and scalability of \themodel on large-scale graphs taking \arxiv as an example. Following the practice in~\citet{bgrl}, we expand the encoder to a 3-layer GCN model. We report the validation and test accuracy of baseline models and ours in Table~\ref{tbl-exp-arxiv}. As demonstrated, many baseline methods cannot run in a full graph manner (on a GPU with 16GB memory). Compared with other scalable methods, 
\themodel can give a competitive performance on \arxiv.

\paragraph{More empirical results and ablation studies.} We conduct further experiments and present the results in Appendix~\ref{appendix:exp} due to the space limit: 1) the performance on heterophilic graphs (Appendix~\ref{appendix:hetero}); 2) the effect of reweighting the importance of positive examples (Appendix~\ref{appendix:pos}); 3) applying the positive example construction strategy to non-contrastive methods (Appendix~\ref{appendix:free}); 4) combining with hard-negative sampling methods (Appendix~\ref{appendix:hard}).

\subsection{Scalability Comparison}\label{sec:exp-scale}
To evaluate the scalability of \themodel on real-world datasets, we compare the total training time and accuracy with state-of-the-art scalable methods BGRL~\citep{bgrl} and CCA-SSG~\citep{cca-ssg} on Arxiv dataset. To further justify the necessity of adopting a large number of negative examples, we additionally adopt GRACE~\citep{grace}, a powerful non-scalable contrastive model. We use sub-sampling strategy to randomly sample a fixed number of negative examples every epoch so that the GRACE model could be fit into a GPU with 16G memory. In Fig.~\ref{fig:scale} we plot the model's performance and the corresponding training time of different methods. We can see that compared with scalable methods CCA-SSG and BGRL, \themodel achieves comparable performance but with the least training time. We can also observe that although reducing the number of negative examples ($k$ in Fig.~\ref{fig:scale}) can enable the model to be trained much faster, the performance drop is significant, which cannot make up for the efficiency benefit. Furthermore, if we continue to use an even larger number of negative examples, GRACE will soon run out of memory. This result demonstrates that \themodel can better balance the training efficiency and model performance.

\subsection{Sensitivity to embedding/projection dimensions}

\begin{wrapfigure}[22]{l}{0.5\textwidth}
\begin{center}
    \includegraphics[width=0.5\textwidth,angle=0]{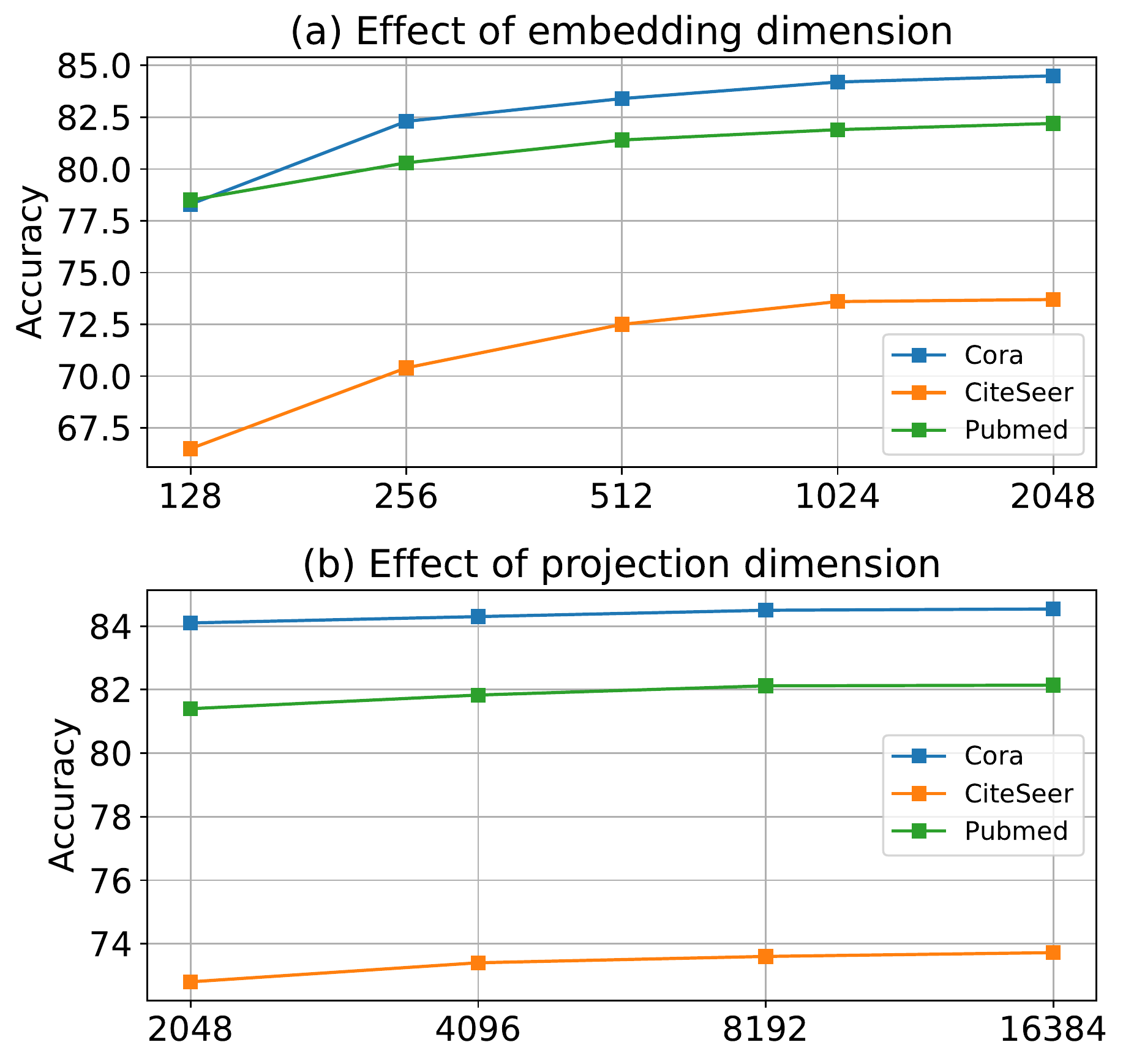}
    \caption{Sensitivity analysis of embedding dimension $d$ and projection dimension $D$.}
    \label{fig:sensitivity}
\end{center}
\end{wrapfigure}

As demonstrated in Theorem~\ref{theorem:error} and Sec.~\ref{sec-negative}, the embedding dimension $d$ and projection dimension $D$ should be two critical hyperparameters affecting the model's performance. In this section, we test \themodel' sensitivities with respect to them in Fig.~\ref{fig:sensitivity}. The observations are two-fold: 1) Similar to other self-supervised learning models~\citep{mvgrl, grace-ad, cca-ssg, afgrl}, \themodel that benefit from a large embedding dimension (e.g., $512$), \themodel's performance also grows as we increase the embedding dimension $d$. However, the optimal embedding dimension for \themodel is even larger, e.g., $2048$ on \cora and \citeseer (note that other methods do not benefit from such a large embedding dimension). This could be justified by our analysis in Theorem~\ref{theorem:error}, which demonstrates that with a larger embedding dimension, the bound of the prediction error will become lower (because the target eigenvalue $\lambda_{d+1}$ becomes larger). 2) Increasing the projection dimension can lead to better but minimal improvement. This indicates that in practice, \themodel can perform quite well without using a huge projection $D$.

\section{Conclusions} 
In this paper, we have presented \themodel, a simple, light-weighted yet effective contrastive model for self-supervised node representation learning. \themodel incorporates two orthogonal techniques to address two key issues in contrastive learning on graph data: reliance on data augmentation and scalability, respectively. The former involves a new definition of positive examples such that the model is free from data augmentation whose design could be more difficult than its counterparts in vision and text, and calls for careful customization for different graph data and tasks. The latter devises an approximated contrastive loss, which reduces the quadratic complexity of traditional contrastive learning to linearity in terms of the graph size. Extensive experimental results on seven public graph benchmarks show its efficiency and efficacy.

{
\bibliographystyle{iclr}
\bibliography{ref}
}

\newpage
\appendix

\section{Proofs}\label{sec-proof}


\subsection{Proof for Theorem~\ref{theorem:error}}\label{appendix-proof-error}
\begin{proof}
    To prove Theorem~\ref{theorem:error}, we first introduce a lemma that shows the optimal representations learned from InfoNCE-like loss function (e.g., Eq.~\ref{eqn:loss-model}) derived from~\citet{contrastive-spectral}:
    \begin{lemma}\label{lemma:eigen}
    Denote the eigendecomposition of the normalized graph adjacency matrix by $ \tilde{\bm{A}} = \bm{U\Lambda} \bm{U}^{\top}$, where $\bm{U}$ and $\bm{\Lambda}$ are the eigenvectors and eigenvalues of $\tilde{\bm{}}$ respectively. Assume node embeddings are free vectors, then a global minimizer of the loss function in Eq.~\ref{eqn:loss-model} is given by:
        \begin{equation}\label{eqn:lemma1}
            \bm{Z}^*  =  (\bm{U}\bm{\Lambda}^{1/2})_{:,1:d},
        \end{equation}
     up to permutations of the eigenvector associated with the same eigenvalue. $1:d$ in Eq.~\ref{eqn:lemma1} indicates the largest $d$ eigenvalues and the associated eigenvectors.
    \end{lemma}

    Besides, ~\citet{contrastive-guarantee} has presented the following theoretical guarantee for the model learned with the matrix factorization loss:
    \begin{lemma}\label{lemma:contrastive-guarantee}
        For a graph $\mathcal{G}$ with symmetric normalized graph adjacency matrix and Laplacian matrix $\tilde{\bm{A}}$ and $\tilde{\bm{L}}$, let $f^*_{mf} \in \mathop{\arg \min}_{f_{mf}: \mathcal{V} \rightarrow \mathbb{R}^d }$ be a minimizer of the matrix factorization loss: $\mathcal{L}_{\rm mf}(\bm{F}) = \Vert  (\bm{I} - \tilde{\bm{L}} - \bm{F}\bm{F}^{\top})\Vert_F^2$, where $\bm{F}$ is the embedding matrix. Then, for any labeling function $\hat{y}: \mathcal{V} \rightarrow [r]$, there exists a linear classifier $\bm{B}^* \in \mathbb{R}^{d \times c}$ with norm  $\Vert B^* \Vert_F  \le 1/(1-\lambda_{d+1})$  such that:
        \begin{equation}
            \mathbb{E}_{v \in \mathcal{V}} \left[\Vert \mathop{y}\limits^{\rightarrow}(v)  - \bm{B}^*f^*(v) \Vert_2^2 \right] \le \frac{1-\phi}{\lambda_{d+1}},
        \end{equation}
        where $\phi$ is the graph homophily ratio defined in Eq.~\ref{eqn:homo-ratio}.
    \end{lemma}
    Then we just need to connect $\bm{Z}^*$ in Eq.~\ref{eqn:lemma1} with the optimal solution of the rthe matrix factorization loss $\mathcal{L}_{\rm mf}(\bm{F}) = \Vert \bm{I} - \tilde{\bm{L}} - \bm{F}\bm{F}^{\top} \Vert_F^2 = \Vert \tilde{\bm{A}} - \bm{F}\bm{F}^{\top} \Vert$. According to Eckart–Young–Mirsky theorem, the optimal $\bm{F}\bm{F}^{\top}$ is $\bm{F}\bm{F}^{\top} = \bm{U}_{:, 1:d}\Lambda_{:, 1:d}\bm{U}_{:, 1:d}^{\top}$  As a result, $\bm{F}^* = \bm{Z}^* = (\bm{U}\Lambda^{1/2})_{:, 1:d}$ is exactly the global minimizer of $\mathcal{L}_{mf}(\bm{F})$. Then the proof is complete.
    
\end{proof}

\subsection{Proof for Theorem~\ref{theorem-rff}}\label{appendix-proof-rff}
\begin{proof}
To prove Theorem~\ref{theorem-rff}, we first introduce the following lemma given by Lemma 1 in~\citep{orf}.
\begin{lemma}\label{lemma-rff}
The Random Fourier Estimation $K_{RFF}(\bm{x},\bm{y})$ is an unbiased estimator of the Gaussian kernel, i.e., $\mathbb{E}_{\bm{\omega}}(\psi(\bm{x})^{\top}\psi(\bm{y})) =  e^{{-\Vert \bm{x} - \bm{y} \Vert^2_2}/{2{\tau}}}$. Let $z = \Vert \bm{x} - \bm{y} \Vert_2 / \sqrt{\tau}$, then the variance of $K_{RFF}(\bm{x},\bm{y})$ is $\mathbb{V}_{\bm{\omega}}(\psi(\bm{x})^{\top}\psi(\bm{y})) = (1 - e^{-z^2})^2 / {2D}$.
\end{lemma}

The lemma shows that the Random Fourier Features can achieve an unbiased approximation for the Gaussian kernel with a quantified variance.

Back to our main theorem, we can derive the following probability using the Chebyshev's inequality:
\begin{equation}\label{eqn-cheb}
    \mathbb{P}(\Delta < {\frac{1-\exp(-4/\tau)}{\sqrt{2D\varepsilon}}}) \ge 1- \frac{\mathbb{V}_{\bm{\omega}}(\psi(\bm{h}_i)^{\top}\psi(\bm{\bm{h}_j}))*2D\varepsilon}{{(1-\exp(-4/\tau))^2}}
\end{equation}
where $\Delta = |{\psi}(\bm{h}_i)^{\top} {\psi}(\bm{h}_j) - \kappa^{G} (\bm{h}_i, \bm{h}_j) |$ denotes the deviation of the kernel approximation. Using the result in Lemma~\ref{lemma-rff}, we can further obtain that the RHS of Eq.~\ref{eqn-cheb} is
\begin{equation}\label{eqn-rhs}
    1 - \frac{(1-e^{-z^2})^2 * \varepsilon}{(1-\exp(-4/\tau))^2}
\end{equation}
As both $\bm{h}_i$ and $\bm{h}_j$ are $\ell_2$-normalized, we have $z^2 = \Vert \bm{h}_i - \bm{h}_j \Vert^2_2 / {\tau} \le 4/\tau$, so we can conclude the stated result:
\begin{equation}
    \mathbb{P}(\Delta < {\frac{1-\exp(-4/\tau)}{\sqrt{2D\varepsilon}}}) \ge 1 - \frac{(1-e^{-z^2})^2 * \varepsilon}{(1-\exp(-4/\tau))^2} \ge 1-\varepsilon
\end{equation}
\end{proof}

\section{Advantages of \themodel over Prior Methods}
We provide a systematic comparison for the proposed \themodel with previous typical methods self-supervised node representation learning, including DGI~\citep{dgi}, MVGRL~\citep{mvgrl}, GRACE~\citep{grace}, GCA~\citep{grace-ad}, BGRL~\citep{bgrl}, CCA-SSG~\citep{cca-ssg} and AFGRL~\citep{afgrl}.

\subsection{Totally augmentation-free with single branch}
Most of previous self-supervised learning models~\citep{dgi, mvgrl, grace, grace-ad, bgrl, cca-ssg} require carefully-defined graph augmentations to obtain positive pairs. However, as noted above, it is hard and costly to design task and dataset-specific graph augmentations. Our method \themodel naturally avoids data augmentation by using the first-order neighborhood information of each node. Moreover, different from all previous methods that have two branches of models (double amounts of inputs, hidden variables, outputs, etc.), our approach only requires a single branch, which makes our method light-weighted. 

\subsection{Projector/Predictor/Discriminator-free}
Most of the previous methods require additional components besides the basic encoder for decent empirical results~\citep{grace, grace-ad}, to break symmetries to avoid trivial solutions~\citep{bgrl, afgrl} or to estimate some score functions in their final objectives~\citep{dgi, mvgrl}. Compared with them, \themodel is much simpler, in both conceptual and practical senses, without using any parameterized model but the basic encoder.

\subsection{Theoretically grounded with linear complexity}
Empirically, InfoNCE-based contrastive methods~\citep{grace, grace-ad} often show better performance than DIM-based methods~\citep{dgi, mvgrl} yet suffer from the scalability issue. Recent non-contrastive methods~\citep{bgrl, afgrl} could maintain decent performance with linear model complexity through asymmetric structures, while the rationale behind their success still remains unclear~\citep{cca-ssg}. As a contrastive model, \themodel is theoretically grounded (through maximizing the mutual information between the target node's embedding and its neighbor's embeddings), and its kernelized approximation of negative loss enables it to scale to large graphs with linear time and memory complexity with respect to the graph size. 

\section{Experiment Details and Additional Empirical Results}\label{appendix-exp}
\subsection{Datasets}\label{appendix-dataset}
The statistics of the used datasets are presented in Table~\ref{tbl-statistics}, and a brief introduction and settings are as follows:
\begin{table}[h]
	\centering
	\caption{Statistics of benchmark datasets.}
	\label{tbl-statistics}
	\small
	\begin{threeparttable}
	{
	\setlength{\tabcolsep}{2.4mm}{
		\begin{tabular}{lcccccc}
			\toprule[1.0pt]
			Dataset  & \#Nodes & \#Edges & \#Classes & \#Features & $\phi$  \\
			\midrule[0.6pt]
            \cora    & 2,708   &  10,556 &  7 & 1,433 & 0.810 \\
            \citeseer & 3,327   &  9,228  & 6  & 3,703 & 0.736 \\
            \pubmed   & 19,717  &  88,651 & 3  & 500 & 0.802 \\
            \cs       & 18,333  &  327,576  & 15  & 6,805 & 0.808  \\
            \physics   & 34,493   &  991,848  & 5  & 8,451 & 0.931 \\
            \computer& 13,752 & 574,418 & 10 & 767 & 0.777  \\
            \photo   & 7,650 & 287,326 & 8  & 745 & 0.827  \\
            \arxiv & 169,343 & 2,332,386 & 40 & 128 & 0.655\\
            \chameleon &  2,277 & 36,101 & 5 & 2,325 & 0.235 \\
            \squirrel & 5,201 & 217,073 & 5 & 2,089 & 0.224 \\
            \actor & 7,600 & 33,544 & 5 & 931 & 0.219 \\     
		\bottomrule[1.0pt]
		\end{tabular}}
	}
	\end{threeparttable}
\end{table}

\cora, \citeseer, \pubmed are three widely used node classification benchmarks~\citep{dataset-cora-citeseer,dataset-pubmed}. Each dataset consists of one citation network, where nodes represent papers and edges represent a citation relationship from one node to another. We use the public split, where each class has fixed $20$ nodes for training, another fixed $500$ nodes and $1000$ nodes for validation/test, respectively for evaluation.

\cs, \physics are co-authorship graphs based on the Microsoft Academic Graph from the KDD Cup 2016 challenge \citep{dataset-coauther}. Nodes are authors that are connected by an edge if they co-authored a paper; node features represent paper keywords for each author’s papers, and class labels indicate the most active fields of study for each author. As there is no public split for these datasets, we randomly split the nodes into train/validation/test (10\%/10\%/80\%) sets.

\computer, \photo are segments of the Amazon co-purchase graph \citep{dataset-amazon}, where nodes represent goods, edges indicate that two goods are frequently bought together; node features are bag-of-words encoded product reviews, and class labels are given by the product category. We also use a 10\%/10\%/80\% split for these two datasets.

For simplicity, we use the processed version of these datasets provided by Deep Graph Library~\citep{dgl}\footnote{\url{https://docs.dgl.ai/en/0.6.x/api/python/dgl.data.html}}. One can easily acquire these datasets using the api provided by DGL.

\arxiv is a directed graph, representing the citation network between all Computer Science (CS) arXiv papers~\citep{ogb}. Each node is an arXiv paper and each directed edge indicates that one paper cites another one. Each paper comes with a 128-dimensional feature vector obtained by averaging the embeddings of words in its title and abstract. The embeddings of individual words are computed by running the skip-gram over the MAG corpus. All papers are also associated with the year that the corresponding paper was published. We use the official split in our experiments.

\chameleon and \squirrel~\citep{musae} are Wikipedia networks introduced. Nodes represent web pages, and edges represent hyperlinks between them. Node features represent several informative nouns in the Wikipedia pages. The task is to predict the average daily traffic of the web page.

\actor is the actor-only induced subgraph of the film-director-actor-writer network used in ~\citet{geom-gcn}. Each node corresponds to an actor, and the edge between two nodes denotes co-occurrence on the same Wikipedia page. Node features correspond to some keywords in the Wikipedia pages. The task is to classify the nodes into five categories in term of words of actor’s Wikipedia.

For \chameleon, \squirrel and \actor, we use the raw data provided by the Geom-GCN~\citep{geom-gcn} paper\footnote{\url{https://github.com/graphdml-uiuc-jlu/geom-gcn}}, and we use the 10-fold split provided.

\begin{table*}[t]
	\centering
	\caption{Details of hyper-parameters of the experimental results in Table~\ref{tbl-exp-main}}
	\label{tbl-para}
	\small
	\begin{threeparttable}
    {   
    {
		\begin{tabular}{lccccccccc}
		    \toprule[1pt]
			 \multirow{2}{*}{Dataset} & \multicolumn{8}{c}{\model} \\
			 \cline{2-9}
			 ~  & Encoder & \# Steps & \# layers & lr & wd &  $d$ & $D$ & $\tau$   \\
			 \hline
			 \cora  & GCN  & 50 & 2 & 5e-4 & 1e-6 & 2048 & 8192 & 0.5 \\
			 \citeseer & GCN & 20 &  1 & 1e-3 & 1e-4 & 2048 & 4096 & 0.5   \\
			 \pubmed & GCN & 50 &  2 & 5e-4 & 0 & 1024 & 8192 & 0.5  \\
			 \computer & GCN & 50 & 2 & 5e-4 & 0 & 1024 & 8192 & 0.8  \\
			 \photo & GCN & 50 & 2 & 5e-4 & 1e-6 & 2048 &  8192 & 0.5  \\
			 \cs & MLP  & 50 &  2 & 5e-4 & 1e-4 & 1024 & 4096 & 0.5  \\
			 \physics & MLP  & 50 &  2 & 5e-4 & 0 & 1024 & 4096 & 0.7 \\
			\bottomrule[1pt]
		\end{tabular}}
	}
	\end{threeparttable}
\end{table*}

\subsection{Hyper-parameters}\label{appendix:hyper}
We provide detailed hyper-parameters on the seven benchmarks in Table~\ref{tbl-para}. All hyper-parameters are selected through a small grid search, and the search space is provided as follows:
\begin{itemize}
    \item Training steps: \{20, 50\}
    \item Number of layers: \{1, 2\}
    \item Embedding dimension $d$: \{512, 1024, 2048\}
    \item Projection dimension $D$: \{2048, 4096, 8192\}
    \item Temperature $\tau$: \{0.5, 0.6, 0.7, 0.8, 0.9, 1.0\}
    \item Learning rate: \{5e-4, 1e-3, 5e-3\}
    \item Weight decay: \{0, 1e-6, 1e-4\}
\end{itemize}
 
\subsection{Extensive empirical studies}\label{appendix:exp}
We further extend \themodel with other designs of positive example construction, negative example selection, and self-supervised objective functions to evaluate the effectiveness of each single component of the proposed method.

\subsubsection{Results on heterophilic graphs}\label{appendix:hetero}
We also investigate the performance on non-homophily graphs, a much more challenging task as directly using first-order neighbors as positive examples without discriminating might be harmful to non-homophily graphs. The results on the three heterophily graphs are presented in Table~\ref{tbl-exp-hete}.
\begin{table*}[t]
	\centering
	\caption{Performance on Heterophily graphs. Results of baseline methods are taken from~\citet{afgcl-guarantee}}
	\label{tbl-exp-hete}
	\small
	\begin{threeparttable}
        \scalebox{1.0}
        {
	\begin{tabular}{lccc}
		\toprule[1.0pt]
             Methods & \chameleon & \squirrel & \actor \\
            \midrule
            DGI  & 60.27$\pm$0.70 & 42.22$\pm$0.63 & 28.30$\pm$0.76\\
            GMI  & 52.81$\pm$0.63 &  35.25$\pm$1.21 & 27.28$\pm$0.87   \\
            MVGRL  & 53.81$\pm$1.09 & 38.75$\pm$1.32 & 32.09$\pm$1.07  \\
            GRACE  & 61.24$\pm$0.53 & 41.09$\pm$0.85 & 28.27$\pm$0.43  \\
            GCA    & 60.94$\pm$0.81 & 41.53$\pm$1.09 & 28.89$\pm$0.50 \\
            BGRL   & 64.86$\pm$0.63 & 46.24$\pm$0.70 & 28.80$\pm$0.54  \\
            AF-GCL & 65.28$\pm$0.53 & 52.10$\pm$0.67 & 28.94$\pm$0.69  \\
            \midrule 
            \themodel & \textbf{68.74}$\pm$\textbf{0.49} & \textbf{52.94}$\pm$\textbf{0.88} & \textbf{33.91}$\pm$\textbf{0.57} \\
            \bottomrule[1.0pt]
	\end{tabular}
		}
	\end{threeparttable}
\end{table*}
Although a little bit counter-intuitive, our method can achieve quite good results on heterophilic graphs. This is probably due to the following reasons: 1) Even though connected nodes may not share the same label, as long as the neighborhood distributions for different classes are different, \themodel is able to recognize the neighborhood patterns of different classes and make the node embeddings for different classes distinguishable. This is also justified in one recent work showing that a GCN model can still perform well on heterophily graphs~\citep{homo-nece}. 2) Data augmentation-based methods tend to keep the low-frequency information while discarding the high-frequency one~\citep{afgcl-guarantee, revisitGCL}, while high-frequency information is much more important for classification on heterophilic graphs~\citep{beyond-hete}. By contrast, edge-wise positive pairs enable \themodel to learn the differences between connected nodes better, thus benefiting heterophilic graphs. 

\subsubsection{Different strategies for constructing positive examples using neighborhoods}\label{appendix:pos}

In addition to treating each neighboring node equally as positive examples, we further explore two other positive sampling strategies: 1) computing the similarities between neighboring nodes and using the most similar one as the positive example (we term it \themodel-max); 2) reweighting the importance of different neighboring nodes according to their similarities, where more similar neighboring node should be assigned a larger weight (we term it \themodel-weight).

For \themodel-max, we first its the nearest neighbor:
\begin{equation}
    \bm{s}_i = \mathop{\arg\min}_{\bm{z}_j, j \in \mathcal{N}(i)} \left\| \frac{\bm{z}_j}{\Vert \bm{z}_j \Vert_2^2} - \frac{\bm{z}_i}{\Vert \bm{z}_i \Vert_2^2} \right\|^2,
\end{equation}
and the objective function is
\begin{equation}
    \mathcal{L}_{LocalGCL-max} = -\frac{1}{|\mathcal{V}|} \sum\limits_{i=1}^{|\mathcal{V}|}  \log \frac{  \exp (\bm{z}_i^{\top}\bm{s}_i/\tau)}{\sum\limits_{k \in \mathcal{V}} \exp (\bm{z}_i^{\top}\bm{z}_k /\tau)} .
\end{equation}

For LocalGCL-weight, we first compute the pair-wise similarities of neighboring nodes:
\begin{equation}
    \text{sim}_{ij} =  \bm{A}_{i,j} \cdot  \frac{\bm{z}_i^{\top}\bm{z}_{j}}{\Vert \bm{z}_i \Vert_2^2 \Vert \bm{z}_j \Vert_2^2}.
\end{equation}

Then, for a target node $i$, it softmax the scores of its neighboring nodes as the weights:
\begin{equation}
    w_{i}(j) = \text{softmax} [\text{sim}_{ij}]_{j \in \mathcal{N}(i)}.
\end{equation}

The objective function of LocalGCL-weight is consequently defined as
\begin{equation}
    \mathcal{L}_{LocalGCL-weight} = -\frac{1}{|\mathcal{V}|} \sum\limits_{i=1}^{|\mathcal{V}|}  \log \frac{\sum\limits_{j \in \mathcal{N}(i)}  w_i(j)\exp (\bm{z}_i^{\top}\bm{z}_j/\tau) / |\mathcal{N}(i)|}{\sum\limits_{k \in \mathcal{V}} \exp (\bm{z}_i^{\top}\bm{z}_k /\tau)}.
\end{equation}

The results of \themodel-max and \themodel-weight, together with the original \themodel is presented in Table~\ref{tbl-abl-pos}:
\begin{table*}[t]
	\centering
	\caption{Performance variation when using different strategies of constructing positive examples from neighboring nodes.}
	\label{tbl-abl-pos}
	\small
	\begin{threeparttable}
    {   
    {
		\begin{tabular}{lccc}
		    \toprule[1pt]
			  & \cora & \citeseer & \pubmed\\
			 \hline
			 \themodel (origin)  & 84.5 $\pm$ 0.4 & 73.6 $\pm$ 0.4 &82.1 $\pm$ 0.5  \\
    \themodel-max & 83.6 $\pm$ 0.5 & 73.7 $\pm$ 0.5 & 81.7 $\pm$ 0.5 \\
    \themodel-weight & 84.9 $\pm$ 0.4 & 73.7 $\pm$ 0.4 & 82.2 $\pm$ 0.5\\
		\bottomrule[1pt]
		\end{tabular}}
	}
	\end{threeparttable}
\end{table*}
From this table, we can observe that using the nearest neighboring node as the only one positive example is likely harmful (except on \citeseer) to the performance, this should be because this operation can discard a lot of useful information in the local neighborhood. Besides, \themodel-weight improves the performance on \cora significantly but makes little difference on \citeseer and \pubmed. This might be because, without supervision, a higher similarity in the latent space does not necessarily indicate a higher probability of sharing the same label. Considering that \themodel-weight requires additional computational cost for computing the weights, and the performance of the original \themodel is already quite good, we just treat every neighboring node equally in this work.

\subsubsection{Combining with negative-sample-free self-supervised learning methods}\label{appendix:free}
using the target node's first-order neighbors as positive examples) could also benefit non-contrastive methods, like BGRL and CCA-SSG. Note that both BGRL and CCA-SSG use the exact same data augmentations methods proposed in GRACE (a contrastive method). As a result, we can simply replace the positive examples in BGRL, and on-diagonal terms in CCA-SSG with the ones used in this paper, thus making them free from data augmentations. In such as a case, the models do not require two-viewed data as inputs, so the size of inputs, intermediate variables, and outputs can be reduced by half. To better demonstrate this point, we further extend our method with BGRL and CCA-SSG.

For BGRL, denote the target embedding of node $i$ by $\bm{h}_i$ and the corresponding prediction by $\bm{z}_i$, our BGRL-Local optimizes the following loss function:
\begin{equation}
   \mathcal{L}_{BGRL-Local} = -\frac{2}{N} \sum\limits_{i \in \mathcal{V}} \sum\limits_{j \in \mathcal{N}(i)} \frac{\bm{z}_j \cdot \bm{h}_i}{\Vert \bm{z}_j \Vert_2^2  \Vert \bm{h}_i \Vert_2^2}
\end{equation}

For CCA-SSG, denote the embedding of node $i$ by $\bm{h}_i$,
 we first compute the local summary of node $i$ by $\bm{z}_i = \frac{1}{|\mathcal{N}(i)|} \sum\limits_{j\in\mathcal{N}(i)}
 \bm{h}_j$. Then we compute two standardized embedding matrices by:
 \begin{equation}
    \tilde{\bm{H}} = \frac{\bm{H} -\mu(\bm{H})}{\sigma(\bm{H})*\sqrt{|\mathcal{V}|}},     \tilde{\bm{Z}} = \frac{\bm{Z} -\mu(\bm{Z})}{\sigma(\bm{Z})*\sqrt{|\mathcal{V}|}}.
 \end{equation}

Finally, CCA-SSG-Local optimizes the following loss function:
\begin{equation}
  \mathcal{L}_{CCA-SSG-Local} = \left\| \tilde{\bm{H}} - \tilde{\bm{Z}} \right\|_F^2 + \lambda \left( \left\|  \tilde{\bm{H}}^{\top}\tilde{\bm{H}} - \bm{I} \right\|_F^2 + \left\|  \tilde{\bm{Z}}^{\top}\tilde{\bm{Z}} - \bm{I} \right\|_F^2 \right).
\end{equation}

We report the performance of BGRL-Local and CCA-SSG-Local compared with the original BGRL and CCA-SSG on CS and Physics, together with their training time/memory cost in Table~\ref{tbl-abl-non-contrastive} (We didn't choose the three citation networks as BGRL performs really bad on them).
\begin{table*}[t]
	\centering
	\caption{Test accuracy, training time cost and memory cost when using the positive pair constructing method in this paper compared with using data augmentation for BGRL and CCA-SSG.}
	\label{tbl-abl-non-contrastive}
	\small
	\begin{threeparttable}
    {   
    {
		\begin{tabular}{lcccccc}
		    \toprule[1pt]
			 \multirow{2}{*}{} & \multicolumn{3}{c}{\cs} & \multicolumn{3}{c}{\physics}\\
			 \cline{2-4} \cline{5-7}
			 ~ & Acc & Time & Memory & Acc & Time & Memory  \\
			 \hline
			 BGRL  & 92.59 $\pm$ 0.17 & 30.4 min &  2.9G & 95.48 $\pm$ 0.08 & 57.7 min & 5.6G  \\
    BGRL-Local & 92.98 $\pm$ 0.23 & 18.2 min &  1.8G & 95.82 $\pm$ 0.11 & 32.3 min & 3.7G  \\
    CCA-SSG & 93.31 $\pm$ 0.22 & 1.1 min &  3.5G & 95.38 $\pm$ 0.06 & 5.9 min & 7.1G  \\
    CCA-SSG-Local & 94.08 $\pm$ 0.19 & 0.7 min &  2.2G & 95.69 $\pm$ 0.09 & 4.3 min & 4.4G \\
		\bottomrule[1pt]
		\end{tabular}}
	}
	\end{threeparttable}
\end{table*}

(BGRL takes much longer time for training, because it does require a lot of training epochs to converge.) As demonstrated in the table. The performance of BGRL and CCA-SSG both get improved after adopting neighboring nodes as positive examples, which highlights the value of graph structure information as self-supervised learning signals. Besides, it also reduces the training time and memory cost because it gets rid of the two-branch architecture caused by data augmentations. 

\subsubsection{Combining with hard negative sampling methods}\label{appendix:hard}
Furthermore, one important advantage of contrastive learning is that it can be incorporated with hard negative sampling techniques~\citep{hardneg-1,hardneg-2,hardneg-3,hardneg-4}. These works either assign weights to different negative samples so that true negative examples are more important than false negative examples, or use mix-up methods to generate even harder negative examples. With these techniques, the performance of contrastive learning can be greatly boosted. To demonstrate this point, we further combine our method with the hard negative mining strategy ProGCL-mix proposed in~\citet{hardneg-4}, and we report the performance comparison on \cora, \citeseer and \pubmed in Table~\ref{tbl-abl-hard}:
\begin{table*}[t]
	\centering
	\caption{Effect of combining \themodel with hard-negative sampling methods.}
	\label{tbl-abl-hard}
	\small
	\begin{threeparttable}
    {   
    {
	\begin{tabular}{lccc}
		\toprule[1pt]
		& \cora & \citeseer & \pubmed\\
			 \hline
			 \themodel (origin)  & 84.5 $\pm$ 0.4 & 73.6 $\pm$ 0.4 &82.1 $\pm$ 0.5  \\
    \themodel+ProGCL-mix & 84.9 $\pm$ 0.5 & 74.2 $\pm$ 0.4 & 82.3 $\pm$ 0.5 \\
		\bottomrule[1pt]
		\end{tabular}}
	}
	\end{threeparttable}
\end{table*}
The improvement is prominent (especially on \citeseer) thanks to the hard-negative sampling strategy(however, ProGCL-mix is not our contribution so we cannot include it in our method in this paper). Besides, we notice that the improvement on \pubmed is smaller than that on the other two datasets, we guess this is because there are only 3 classes of nodes on Pubmed, so it is less effective to mine hard negatives than 7 classes on \cora and 6 on \citeseer. 
\end{document}